\documentclass{article}
\usepackage{spconf,amsmath,graphicx}
\usepackage{times}
\usepackage{epsfig}
\usepackage{graphicx}
\usepackage{amsmath}
\usepackage{amssymb}
\usepackage{multirow}
\usepackage[ruled]{algorithm2e}
\usepackage[pagebackref=false,breaklinks=true,letterpaper=true,colorlinks,bookmarks=false]{hyperref}

\usepackage{subcaption}
\usepackage{algorithmic}
\usepackage{paralist}

\def\x{{\mathbf x}}

\def \x {\mathbf{x}}
\def \N {\mathcal{N}}

\title{Graph Convolution for Re-ranking in Person Re-identification}
\name{
Yuqi Zhang$^{\star}$ \quad
Qian Qi$^{\star \mathsection}$ \thanks{$\mathsection$ Equal contribution} \quad
Chong Liu$^{\dagger \ddagger}\thanks{The work was done when Chong Liu was intern at Alibaba Group}$ \quad
Weihua Chen$^{\star}$ \quad
Fan Wang$^{\star}$ \quad
Hao Li$^{\star}$ \quad 
Rong Jin$^{\star}$ \qquad}

  \address{$^{\dagger}$ State Key Laboratory of Computer Science, Institute of Software, Chinese Academy of Sciences \\
      $^{\ddagger}$ University of Chinese Academy of Sciences, Beijing China\\
      $^{\star}$ Machine Intelligence Technology Lab, Alibaba Group\\}

\begin{document}
%\ninept
%
\maketitle
\begin{abstract}
Nowadays, deep learning is widely applied to extract features for similarity computation in person re-identification (re-ID). However, the difference between the training data and testing data makes the performance of learned feature degraded during testing.  Hence, re-ranking is proposed to mitigate this issue and various algorithms have been developed. However, most of existing re-ranking methods focus on replacing the Euclidean distance with sophisticated distance metrics, which are not friendly to downstream tasks and hard to be used for fast retrieval of massive data in real applications. In this work, we propose a graph-based re-ranking method to improve learned features while still keeping Euclidean distance as the similarity metric. Inspired by graph convolution networks, we develop an operator to propagate features over an appropriate graph. Since graph is the essential key for the propagation, two important criteria are considered for designing the graph, and different graphs are explored accordingly. Furthermore, a simple yet effective method is proposed to generate a profile vector for each tracklet in videos, which helps extend our method to video re-ID. Extensive experiments on three benchmark data sets, e.g., Market-1501, Duke, and MARS, demonstrate the effectiveness of our proposed approach.
\end{abstract}
%
%%%%%%%%% BODY TEXT
\section{Introduction}
%introduce re-id
Person re-identification (re-ID) aims to retrieve images of the same person from the gallery set given a query image~\cite{zhong2017re}. A standard pipeline is to extract features for images in both the gallery set and the query based on a pre-trained deep model, and then return the top-ranked images in the gallery, where the similarity is measured by the Euclidean distance~\cite{zheng2017discriminatively}. However, due to the difference between the distribution of the training set from the deep model and that of the testing set, directly generating features based on the pre-trained model may result in a sub-optimal performance. Many post-process methods have been proposed to mitigate the challenge while re-ranking is one of the most effective approaches for outstanding performance~\cite{bai2019re,saquib2018pose,zhong2017re}.

\begin{figure}
\centering
\includegraphics[width=3.2in]{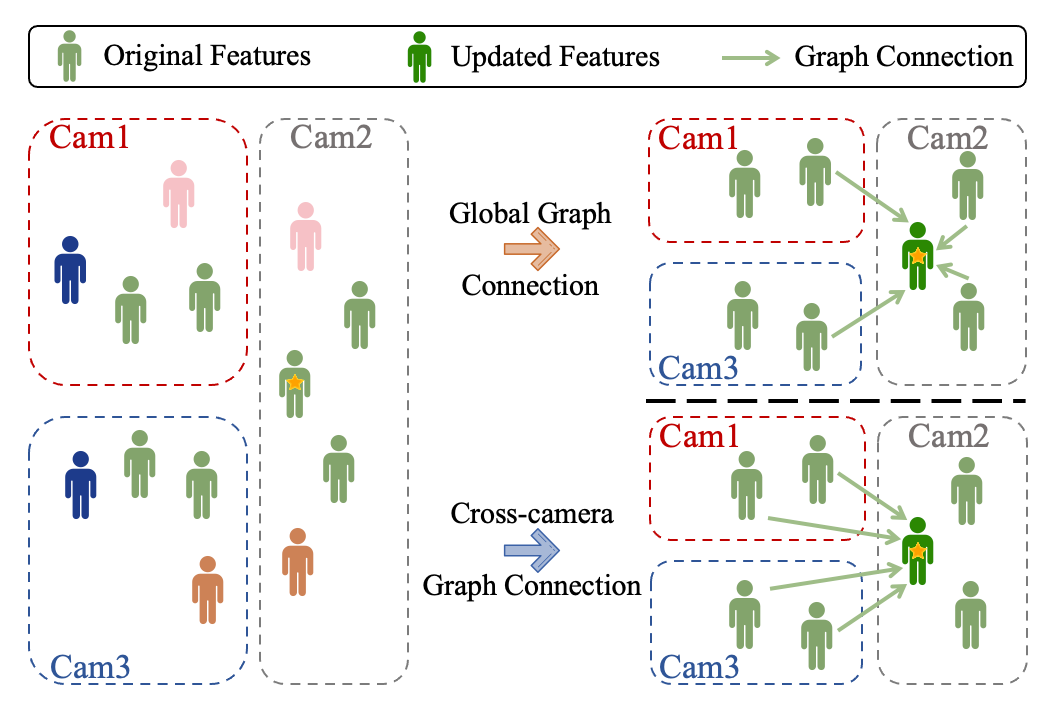}
\caption{Illustration of graphs with two proposed criteria. The person with the star denotes the target image and the arrows indicate its $k$-nearest neighbors. People with the same color hold the same ID. Corresponding to the two criteria, we generate two graphs (i.e., Global graph: connecting the $k$-nearest neighbors in all cameras, and Cross-camera graph: connecting the $k$-nearest neighbors from different cameras of the target person, excluding those from the same camera).\label{fig:illu}}
\end{figure}

%introduce re-ranking
Given features from the deep model, re-ranking is to recalculate the similarity of images by introducing other information and use sophisticated similarity metrics~\cite{bai2017scalable, bai2019re, bai2017ensemble, saquib2018pose,yu2017divide,zhong2017re} to rearrange the ranking list. Current SOTA methods $k$-reciprocal encoding~\cite{zhong2017re} or ECN~\cite{saquib2018pose}can surpass the performance of original features by a large margin. Despite the success, the sophisticated distance metrics adopted by these re-ranking methods are much more complicated than Euclidean distance, which are not friendly to downstream tasks and hard to be used for fast retrieval of massive data in real applications.
Therefore, some work~\cite{luo2019spectral} tries to optimize the original features based on Euclidean distance. But their performance still cannot catch up with $k$-reciprocal encoding.

%introduce our work
Instead of figuring out an appropriate and sophisticated distance metric, in this work, we aim to modify the original features while Euclidean distance can still be directly used as the similarity measure. Inspired by graph convolution networks (GCN)~\cite{kipf2016semi}, we adopt the graph convolution operator to propagate features over a graph, so as to improve the representation of each image. More specifically, we construct our graphs for feature propagation with two criteria. First, the changes in features should be moderate after re-ranking to preserve the knowledge learned in the pre-trained feature representation model. Therefore, only features from nearest neighbors can be propagated to the target image. This criterion essentially shares a similar idea with other successful re-ranking methods~\cite{zhong2017re,saquib2018pose}. Second, features propagated from different cameras should be emphasized. This criterion has been rarely investigated but it is helpful to eliminate the bias from cameras. With these criteria, we develop a feature propagation method that obtains features from two graphs simultaneously. 

Fig.~\ref{fig:illu} illustrates the proposed graphs with our two criteria. Both of two graphs take the $k$-nearest neighbors into account for each image. The difference is that in the global graph, the $k$-nearest neighbors of each image are from all cameras, while in the cross-camera graph, the $k$-nearest neighbors are from only different cameras of a given image. Then, we apply a graph convolution operator on these two graphs. After obtaining propagated features from two graphs, their weighted combination is treated as the final feature representation to re-compute the ranking list based on Euclidean distance. To the best of our knowledge, this is the first work that achieves state-of-the-art performance in re-ranking with Euclidean distance.

The main contributions of our work can be summarized as follows.
\begin{compactitem}
\item We propose the criteria of feature propagation for re-ranking and develop a graph convolution based re-ranking (GCR) method accordingly. The features obtained from our method are still in the Euclidean space, which can be easily used in downstream tasks and available for fast retrieval of massive data in real applications.
\item Along with the GCR, to take full advantage of multi-frame information in video re-ID task, we further present a simple yet effective method to generate a profile vector for each tracklet in video re-ID, called profile vector generation (PVG).
\item As the image-level re-ID task can be considered as a video re-ID with only one image in each tracklet, we combine GCR and PVG together to build our final solution, \textit{i.e.} Graph Convolution Re-ranking for Video (GCRV), which achieves state-of-the-art performance on the ReID benchmarks in both image-level and video-level re-ID tasks.
\end{compactitem}

% The rest of this paper is organized as follows. Section~\ref{sec:related} reviews the related work. Section~\ref{sec:method} analyzes feature propagation for re-ranking and proposes our graph convolution based re-ranking method. Section~\ref{sec:pvg} presents our method on how to generate profile vectors to represent tracklets for video re-ID. Section~\ref{sec:exp} provides the experiment analysis of our methods on the benchmark data sets. Section~\ref{sec:conclude} concludes this work with the future direction.

\begin{figure*}
\centering
\includegraphics[width=1.0\textwidth]{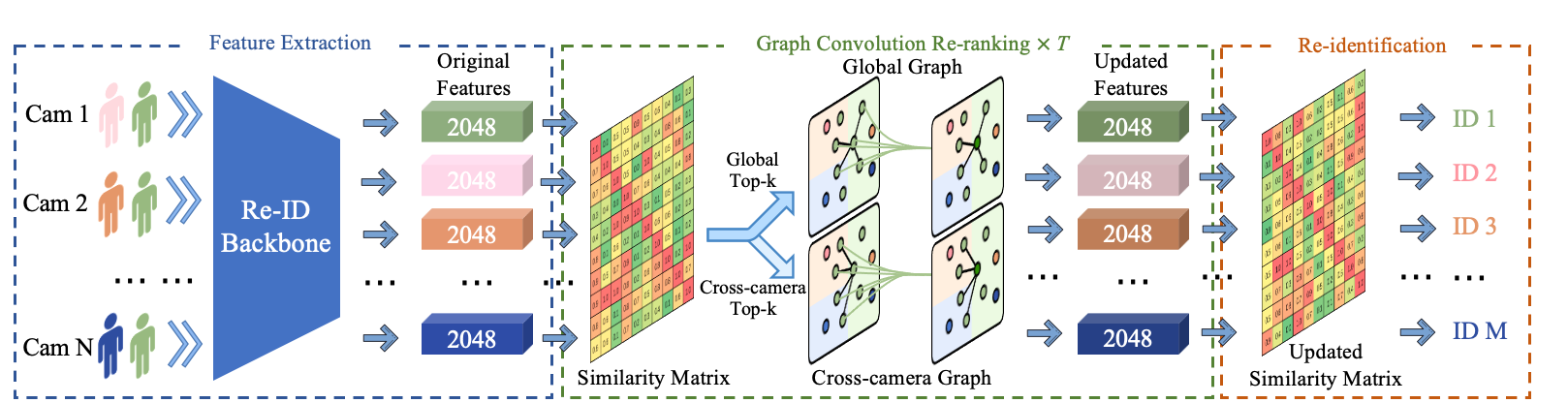}
\caption{The pipeline of the proposed graph convolution based re-ranking (GCR) method.}
\label{fig:framework}
\end{figure*}

\section{Graph Convolution For Re-ranking}
\label{sec:method}

% Different from previous re-ranking methods, which compute similarity with the sophisticated distance metric rather than Euclidean distance, we aim to improve representations while simply using Euclidean distance for retrieval. These features can be more flexible for downstream tasks. %It is worth noting that the improved representations could be served for both image-based re-ID and video-based re-ID.    % last sentence added by kugang

% The key challenge in the proposed method is to have the appropriate criterion for generating new features. Considering that the provided features are well trained, the changes in the features should be mild. Besides, features of the same person can be from different cameras and it is important to align features across multiple cameras to eliminate the bias from cameras. Therefore, we propose to propagate features over a graph with following criteria.
We propose to propagate features over a graph with following criteria.

\begin{enumerate}
\item Given an image, only features from its $k$-nearest neighbors should be propagated.
\item Nearest neighbors from different cameras should be emphasized.
\end{enumerate}
The first criterion implies a sparse graph which tries to mitigate the noisy features by taking their neighbors into account. The second criterion is to align features from different cameras, which is rarely investigated and important for reducing the gap between training and testing data. In the following sections, we will illustrate the details of our graph convolution based re-ranking (GCR) method, especially how to build graphs with these two criteria.

%We note that most of existing methods~\cite{saquib2018pose,zhong2017re} have already encoded the first criterion that leverages the knowledge only from nearest neighbors while the second criterion is rarely investigated, which is important for domain adaptation.

\subsection{$k$-Nearest Cross-camera Graph}
\label{two_graph}

Considering the first proposed criterion, we propose a global graph first. To make sure that there are samples from different cameras for propagation, which is suggested in the second criterion, we also introduce an cross-camera graph with $k$-nearest neighbors from different cameras as follows.

\begin{enumerate}
\item For the $i$-th image, obtain its $k$-nearest neighbors $\N_i^{\mathrm{diff}:k}$ from different cameras with the original features.
\item For the $i$-th row of $A$, we compute the similarity as
\begin{eqnarray}
A_{i,j} = \left\{\begin{array}{cc}\exp(-\|\x_i - \x_j\|_2^2/\gamma)&j\in\N_i^{\mathrm{diff}:k}\\1&j=i\\0&o.w.\end{array}\right.
\end{eqnarray}
\end{enumerate}
We denote the resulting similarity matrix as $A_{nonsym}^{cross}$, which is the similarity matrix across different cameras. Note that we include the $i$-th image itself in the similarity graph to calibrate the feature after propagation and make it comparable to the one from the global propagation. 

Propagation with the cross-camera graph emphasizes the relationship between the image and its $k$-nearest neighbors from different cameras. It helps to eliminate the bias from cameras in the similarity matrix and align features across multiple cameras.  
With two obtained similarity matrices, we have our final propagation criterion as
\begin{eqnarray}\label{eq:criterion}
\begin{aligned}
\widetilde{X} = &\alpha D_{row:global}^{-\frac{1}{2}} A_{nonsym}^{global} D_{col:global}^{-\frac{1}{2}} X +\\
                &(1-\alpha) D_{row:cross}^{-\frac{1}{2}} A_{nonsym}^{cross} D_{col:cross}^{-\frac{1}{2}} X
\end{aligned}
\end{eqnarray}
where $\alpha$ is the parameter to balance the weights between two propagation procedures. Note that the parameter $k$ can be different when generating these two similarity matrix, we denote them as $k_g$ and $k_c$, respectively. Finally, the obtained features can be iteratively updated with the same criterion in Eq.\ref{eq:criterion} as
\begin{eqnarray}\label{eq:nonsym}
\begin{aligned}
X_{t+1} = &\alpha D_{row:global}^{-\frac{1}{2}} A_{nonsym}^{global} D_{col:global}^{-\frac{1}{2}} X_t + \\
          &(1-\alpha) D_{row:cross}^{-\frac{1}{2}} A_{nonsym}^{cross} D_{col:cross}^{-\frac{1}{2}} X_t
\end{aligned}
\end{eqnarray}
where $t$ indicates the iteration index, from $1$ to $T$. $T$ is the total number of iterations and $X_1 = X$. The similarity matrices $A_{nonsym}^{global}$ and $A_{nonsym}^{cross}$ change during iterations. The whole pipeline is shown in Fig.~\ref{fig:framework}.

%An example with real images can be found in Fig.~\ref{fig:real}
%\begin{figure}[!ht]
%\centering
%\includegraphics[width=3.2in]{framework}
%\caption{Illustration of graph convolution for re-ranking with real %images.\label{fig:real}}
%\end{figure}

\section{Profile Vector Generation for Video Re-ID}
\label{sec:pvg}
Besides re-ranking for images, its application for video re-ID attracted much attention recently.It's important to take full advantage of these multiple images in the tracklet to build a robust feature vector of this tracklet. Therefore, we propose a profile vector generation (PVG) method to extract a profile vector for each tracklet. And then our GCR method from image-level re-ID task can be extended to be applied in the video re-ID task.

In this paper, we expect the new profile vector $\hat{\x}_c$ of the $c$-th tracklet should be near to the features of images in the the $c$-th tracklet, and meanwhile far away from the other features in the same camera. Hence, a ridge regression is involved to achieve this constraint. For each $\hat{\x}_c$, the optimization problem becomes
\begin{eqnarray}
\min_{\hat{\x}_c} \frac{1}{n_z} \sum_{i=1}^{n} (\x_i^\top \hat{\x}_c - z_i^c)^2+\frac{\lambda_p}{2}\|\hat{\x}_c\|_2^2
\end{eqnarray}

where $n_z$ is the total number of images in the $z$-th camera, and the $z_i^c$ is the binary label whether the feature $\x_i$ comes from the $c$-th tracklet. The $\|\hat{\x}_c\|_2$ is a regularization term. 
For each tracklet, the profile vector can be calculated with the closed-form solution as
\begin{small}
\begin{eqnarray}\label{eq:profile}
\hat{\x}_c = norm\big((X_z^\top X_z+n_z\lambda_p I)^{-1}(\frac{1}{n_z^c}\sum_{i:y_i=c}x_i - \frac{1}{n_z}\sum_{i=1}^{n_z} x_i)\big)
\end{eqnarray}
\end{small}
where $I$ is the identity matrix and $X_z$ consists of all images from the $z$-th camera. $norm(\cdot)$ is a l2-norm operator. Compared with the mean vector, the profile in Eq.~\ref{eq:profile} eliminates the mean vector $\frac{1}{n_z}\sum_{i=1}^{n_z} x_i$ of images from the same camera to reduce the bias from different cameras and leverages the geometric information from the covariance matrix $X_z^\top X_z$.

Although designed for video-based re-ID, the profile vector is also available for image-based re-ID, where each image could be viewed as a tracklet with only one frame.

\begin{table*}[!ht]
\begin{center}
\begin{tabular}{|c|c|cc|cc|cc|}
\hline
\multirow{2}{*}{Method} & \multirow{2}{*}{Reference} &  \multicolumn{2}{c}{Market}	& \multicolumn{2}{|c}{Duke}	& \multicolumn{2}{|c|}{MARS}       \\    
	&					&  Rank-1 & mAP 			 &  Rank-1 & mAP 				&  Rank-1 & mAP \\
\hline
ISP~\cite{zhu2020identity}              & ECCV20    & 95.3 & 88.6   & 89.6 & 80.0 	& -    & - 	  \\
MPN~\cite{ding2020multi}                & TPAMI20   & 96.3 & 89.4   & 91.5 & 82.0   & -    & - 	  \\
% CircleLoss~\cite{sun2020circle}         & CVPR20    & 96.1 & 87.4   & -    & -      & -	   & -    \\
%MGN+CircleLoss~\cite{sun2020circle}     & CVPR20    & 96.1 & 87.4   & -    & -      & -	   & -    \\
%Resnet50+CircleLoss~\cite{sun2020circle} & CVPR20   & 94.2 & 84.9   & -    & -      & -    & - 	  \\
% AGRL~\cite{wu2020adaptive}              & TIP20    & -    & -      & -    & -      & 89.5 & 81.9 \\
% VKD~\cite{porrello2020robust}           & ECCV20    & -    & -      & -    & -      & 89.4 & 83.1 \\	
MGH~\cite{yan2020learning}              & CVPR20    & -    & -      & -    & -      & 90.0 & 85.8 \\
\hline
SOTA features 				& CVPR20    & 96.3 & 89.4   & 91.5 & 82.0   & 90.0 & 85.8 \\
SOTA+KR~\cite{zhong2017re}  & CVPR17 	& 95.6 & 94.5   & 90.5 & 89.6 	& 88.8 & 90.7 \\
SOTA+ECN~\cite{saquib2018pose}   & CVPR18    & 95.1 & 94.0 	& 90.8 & 88.3 	& 92.7 & 90.5 \\
SOTA+LBR~\cite{luo2019spectral} 		& ICCV19	& 95.0 & 92.3 	& 89.7 & 85.8	& 91.4 & 87.5 \\
SOTA+GCRV 				    & -	& $\textbf{96.6}$ &  $\textbf{95.1}$ & $\textbf{92.9}$ &  $\textbf{91.3}$				&  $\textbf{93.8}$ &  $\textbf{92.8}$  \\
%SOTA+GCRV+KR 			    & -	& $\textbf{96.9}$ &  $\textbf{95.2}$ & $\textbf{93.4}$ &  $\textbf{91.5}$				&  $\textbf{94.3}$ &  $\textbf{93.1}$  \\
\hline
\end{tabular} 
\end{center}
\caption{Comparison with state-of-the-art methods on Market-1501, Duke and MARS. %Results marked with $\ast$ are based on the same baseline from~\cite{luo2019bag}. 
The \textbf{bold} indicates the best performance.\label{tab:comparison}}
\end{table*}

\section{Experiments}
\label{sec:exp}

\subsection{Datasets}

In our experiments, we evaluate the proposed GCR on both image-based including Market-1501~\cite{zheng2015scalable} and Duke-MTMC-re-ID (Duke)~\cite{ristani2016performance}, and video-based re-ID data sets, \textit{e.g.} MARS~\cite{zheng2016mars}. 

\noindent\textbf{Market-1501}~\cite{zheng2015scalable} is a widely-used benchmark for person re-id with $1,501$ identities from 6 cameras in total $750$ identities ($12,936$ images) are used for training, $751$ identities ($19,732$ images) are used for testing. %Although high accuracy has been achieved in recent years, we demonstrate that the proposed re-ranking method can further boost the performance.

\noindent\textbf{Duke-MTMC-re-ID (Duke)}~\cite{ristani2016performance} dataset consists of $1,812$ people from 8 cameras. Training and test sets both consist of $702$ persons. %The training set includes $16,522$ images, the gallery $17,661$ images, and the query set $2,228$ image. Person bounding boxes in this dataset are manually annotated.

\noindent\textbf{MARS}~\cite{zheng2016mars} is used as a large-scale video-based person re-ID datasets in our experiments. It consists of $17,503$ tracks and $1,261$ identities. %Each track has $59$ frames on average. Similar to image-based Market-1501, MARS includes $3,248$ distractor tracks to make the person re-ID more challenging.

% \subsection{Implementation Details}
% We extract re-ID features with models from strong baseline~\cite{liu2019spatially, luo2019bag, luo2019strong}. \cite{luo2019bag} provides public pre-trained models for Market and Duke while \cite{liu2019spatially} provides models for MARS. All these baseline models achieve competitive performance and the dimension of features is $2,048$. As current re-ranking methods always provide their results on different baselines and thus hard to compare with each other, we use an open-source state-of-the-art baseline and hope other researchers could be convenient to compare results with ours under the same baseline in future.

\subsection{Comparison with State-of-the-Art Methods}
\label{sec: state_of_the_art}
Table~\ref{tab:comparison} compares the proposed method to state-of-the-art re-ranking methods. 
%We set $k_{g}$=15, $k_{c}$=3 for all three datasets. 
To make a fair comparison, we reproduce the results of the most commonly used re-ranking methods under the same features. The proposed method outperforms reranking methods KR, ECN and LBR by a large margin.
%Also GCRV method could be further improved if combined with KR.
It is worth noticing that after re-ranking with our GCRV, the feature is still in the Euclidean space which can be easily used in downstream tasks and available for fast retrieval of massive data in real applications.

\subsection{Ablation Study}
% \subsubsection{Effect of Global and Cross-camera Graphs}

\begin{figure}[t]
\centering
\begin{subfigure}[b]{0.475\linewidth}
\centering 
\includegraphics[width=\textwidth]{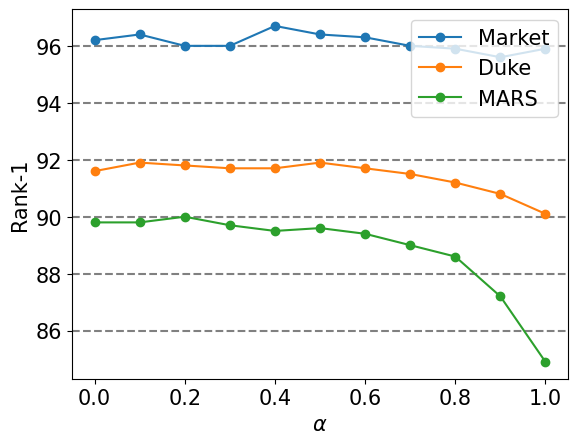}
\caption{rank-1 curve}
\end{subfigure}
\begin{subfigure}[b]{0.475\linewidth}
\centering                                                          
\includegraphics[width=\textwidth]{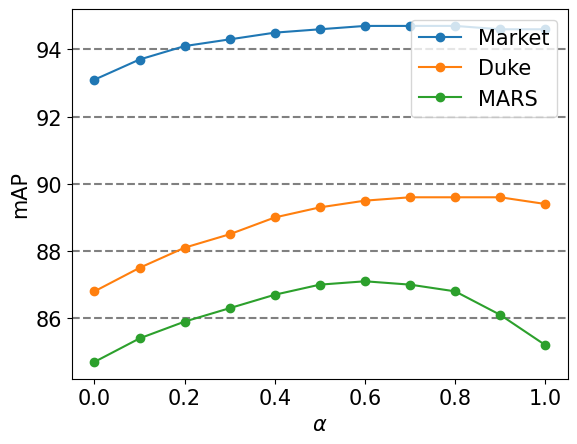}        
\caption{mAP curve}
\end{subfigure}
\caption{The performance curve under different $\alpha$.}% for two different graphs.}
\label{fig:alpha_tuning} 
\end{figure}

To make a fair comparison, we use BoT~\cite{luo2019bag} features in the ablation study. The trade-off hyper-parameter between two graphs is fixed as $\alpha = 0.7$. We plot accuracy curves with respect to the different $\alpha$ in Fig.~\ref{fig:alpha_tuning}. Rank-1 saturates for $\alpha < 0.7$ while mAP reaches the peak at $\alpha = 0.7$. Since mAP is often more important for retrieval cases, we select the hyper-parameter for the sake of better mAP.

\begin{table}[t]
\begin{center}
%\footnotesize
\resizebox{0.95\linewidth}{!}{
\begin{tabular}{|c|cc|cc|cc|}
\hline
\multirow{2}{*}{Method} 	&  \multicolumn{2}{|c}{Market}   & \multicolumn{2}{|c}{Duke} 		& \multicolumn{2}{|c|}{MARS} \\
  				&  Rank-1 & mAP 						&  Rank-1 & mAP 						&  Rank-1 & mAP \\
\hline
baseline 	& 94.5      & 	85.9 						& 86.5 & 76.4 							& 85.8 & 79.7 \\
+GCR  		& 96.0 	    &	94.7						& 91.5 & $\textbf{89.6}$ 							& 86.6 & 85.3 \\
+PVG   		& 94.6      &   86.3 						& 86.9 & 76.0 							& 88.6 & 80.6 \\
+GCRV       & $\textbf{96.1}$  & $\textbf{94.7}$		& $\textbf{91.6}$ & 89.2 				& $\textbf{89.0}$ & $\textbf{87.0}$ \\

% in fact Market GCRV is 96.14  94.71 and GCR is 96.05  94.69
\hline
\end{tabular}}
\end{center}
\caption{Comparison of GCR, PVG and GCRV on Market-1501, Duke and MARS.\label{tab:ablation}}
\end{table}

Then, we incorporate PVG to GCR and compare the performance of GCR and GCRV in Table~\ref{tab:ablation}. It is not surprising to observe that GCR achieves dramatic improvement on different data sets compared to the baseline. It is because re-ranking can effectively mitigate the challenge from different cameras. On the image-based re-ID, GCRV achieves similar result with GCR. But on the video-based re-ID dataset MARS, GCRV demonstrates a better performance than GCR. It confirms that GCRV is more appropriate for the video-based re-ID. 
%better on Market but a little bit worse on Duke. It is because PVG is designed for a set of multiple images, the variant for a single image should be further investigated.

\subsection{Efficiency}
\label{efficiency}
%In real-world applications where the gallery size is very large, traditional re-ranking methods such as k-reciprocal may suffer from heavy computation. 
Table~\ref{tab:speed} lists the computation time of different re-ranking methods on the same Market-1501 dataset with the same hardware settings of 24 cores Platinum 8163 CPU. The similarity matrix size is 3368 queries * 15913 galleries, and our time complexity is $\mathcal{O}\left(N^{2} \log N\right)$. As can be seen, K-reciprocal (KR) and ECN suffer from low computation speed due to complex set operations.
%Although LBR is fast, its accuracy is limited as shown in Table~\ref{tab:comparison}
On the other hand, the proposed method relies only on simple matrix operations and achieves better efficiency.

\begin{table}[t]
\begin{center}
\begin{tabular}{|c|c|c|c|}
\hline
Method	& KR~\cite{zhong2017re}	&ECN~\cite{saquib2018pose}   & proposed\\
\hline
Time(s) & 76 & 72 & 24 \\

\hline
\end{tabular}
\end{center}
\caption{The computation time of re-ranking methods on Market-1501.}
\label{tab:speed}
\end{table}

\section{Conclusion}
\label{sec:conclude}
In this paper we propose a graph convolution based re-ranking method for person re-ID. Unlike previous methods, we propose to learn features with propagation over graphs and re-compute similarity with the standard Euclidean distance. By investigating the criteria for propagation, we develop different similarity graphs and propagate features from both graphs for a single image. Empirical study with strong baseline verifies the effectiveness of the proposed method. 

In our method, the convolution parameter of $W$ is set to be an identity matrix. With a small set of labeled images from the target domain, we can improve the re-ranking method with a learnable $W$. Applying our method for semi-supervised re-ranking can be our future work.

\clearpage

\bibliographystyle{IEEEbib}
\bibliography{egbib}

\end{document}